\pdfoutput=1
\documentclass[11pt]{article}

\usepackage{acl}

\usepackage{times}
\usepackage{latexsym}

\usepackage[T1]{fontenc}
\usepackage[utf8]{inputenc}

\usepackage{microtype}

\usepackage{graphicx}
\usepackage{cleveref}
\usepackage{pgfplots}
\usepackage{caption}
\usepackage{subcaption}
\usepackage{bm}
\usepackage{fontawesome}
\usepackage{siunitx}
\usepackage{booktabs}
\usepackage{adjustbox}

\title{How to Adapt Pre-trained Vision-and-Language Models to a Text-only Input?}

\author{Lovisa Hagström\textsuperscript{\normalfont1}\hspace{5mm}Richard Johansson\textsuperscript{\normalfont1,2} \\
  \textsuperscript{1}Chalmers University of Technology, 
  \textsuperscript{2}University of Gothenburg\\
  \texttt{\{lovhag, richajo\}@chalmers.se}}

\begin{document}
\maketitle
\begin{abstract}
Current language models have been criticised for learning language from text alone without connection between words and their meaning. Consequently, multimodal training has been proposed as a way for creating models with better language understanding by providing the lacking connection. We focus on pre-trained multimodal vision-and-language (VL) models for which there already are some results on their language understanding capabilities. An unresolved issue with evaluating the linguistic skills of these models, however, is that there is no established method for adapting them to text-only input without out-of-distribution uncertainty. To find the best approach, we investigate and compare seven possible methods for adapting three different pre-trained VL models to text-only input. Our evaluations on both GLUE and Visual Property Norms (VPN) show that care should be put into adapting VL models to zero-shot text-only tasks, while the models are less sensitive to how we adapt them to non-zero-shot tasks. We also find that the adaptation methods perform differently for different models and that unimodal model counterparts perform on par with the VL models regardless of adaptation, indicating that current VL models do not necessarily gain better language understanding from their multimodal training. 
\end{abstract}

% WORDS TO CONSIDER
% pre-trained vs pre-trained? DONE
% fine-tune vs finetune? DONE 
% use text-only version!

\section{Introduction}
Having models learn language from text alone has been criticised based on several aspects, from fundamental arguments about how language works \citep{bender-koller-2020-climbing} to findings on lack of certain information in text \citep{gordon2013reporting, paik-etal-2021-world}. To train language models on more sources than text is therefore a proposed direction for creating language models with better language understanding \citep{bisk-etal-2020-experience}. These models would then become multimodal, with the capability to process both text and information from other modalities. 

\begin{figure}[t]
    \centering
    \includegraphics[width=\linewidth]{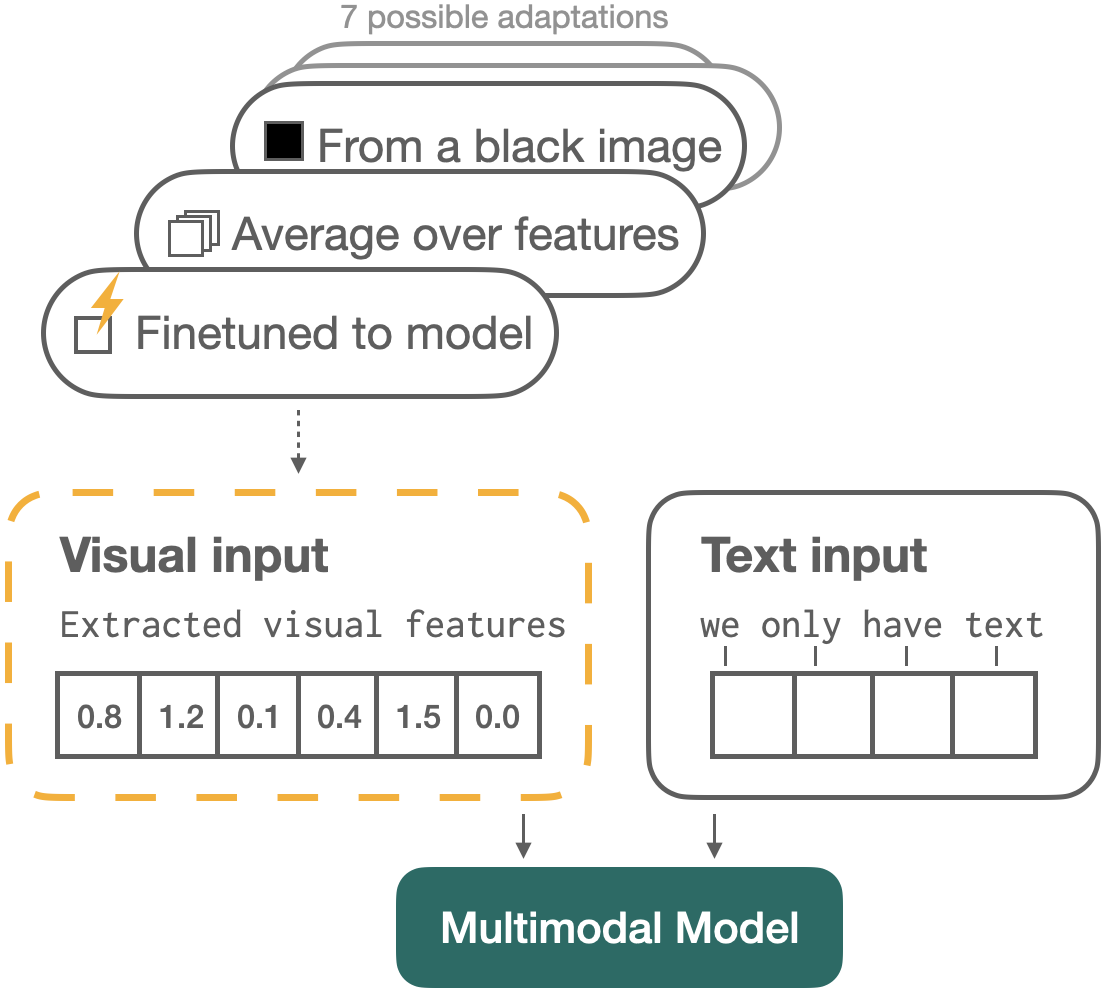}
    \caption{An overview of different ways to adapt a multimodal model to text-only input. It showcases three of the seven adaptations evaluated in this work.}
    \label{fig:overview}
\end{figure}
The multimodal models of interest in this work are vision-and-language (VL) models that have been trained on images and their corresponding captions or visual questions \citep{NEURIPS2019_c74d97b0,tan-bansal-2019-lxmert,Su2020VL-BERT:,visualbert,chen2020uniter}. These models are performant on several image-text tasks such as image captioning and VQA, while there also is an increased interest for evaluating how their natural language understanding is influenced by their multimodal training \citep{iki-aizawa-2021-effect,yun-etal-2021-vision-language}. 

It is however tricky to investigate the pure natural language understanding of the aforementioned VL models, since their language processing is conditioned on visual features. For certain investigations, we may simply wish to evaluate the models on text-only domains, while these models have not been developed for this purpose. If we do not attend to the issue of accurately adapting VL models to text-only domains we risk evaluating them out-of-distribution and fail to accurately measure their natural language understanding capabilities.

Different methods for adapting VL models to a text-only input have already been tried and we have some results on the natural language understanding capabilities of these models \citep{iki-aizawa-2021-effect,yun-etal-2021-vision-language}. %However, it has not been ascertained whether the methods used to adapt the models to the out of distribution text-only input worked as intended. %and we do not know if the models were functioning adequately during evaluation. 
%Consequently, we do not know if these results correctly measured the language understanding of the models, or if they also measured the fit of the adaptation to a unimodal input.
%Work has been performed on evaluating the language understanding of pre-trained VL models . Consequently, different methods for adapting these models to a unimodal input have already been tried and we have some results on the natural language understanding capabilities of these models. However, we do not know if the models were functioning adequately during evaluation, since it has not been possible to ascertain whether the methods for adapting the models to the out of distribution text-only input worked as intended. Therefore, we do not know if these evaluations correctly measured the language understanding of the models, or if they also measured the fit of the adaptation to a unimodal input.
However, no systematic search for the best way to adapt VL models to a text-only input has been performed and it is unclear how well the VL models work with the %already tried
previously proposed
adaptations. If we wish to continue the search for better natural language understanding in multimodal models, 
%continue the investigation of the natural language understanding capabilities of pre-trained VL models, 
we should ensure that we evaluate them in the best way possible. In this work, we search for the best method for adapting existing VL models to a text-only input, as illustrated in \Cref{fig:overview}.\footnote{Code available at \url{https://github.com/lovhag/adapt-pre-trained-VL-models-to-text}}

With the adaptations in place, we can then compare the VL models to their unimodal text-only counterparts. This will complement already existing results on the natural language understanding capabilities of VL models and the effect of multimodal training.

The contributions of our work are as follows:

\begin{itemize}\addtolength{\itemsep}{-0.5\baselineskip}
    \item We investigate and compare seven methods for adapting LXMERT \citep{tan-bansal-2019-lxmert}, VisualBERT \citep{visualbert} and CLIP-BERT \citep{norlund-etal-2021-transferring} to a text-only input (\Cref{sec:adaptations}). Two of these adaptations have already been used in previous investigations of the linguistic capabilities of VL models \citep{frank-etal-2021-vision,iki-aizawa-2021-effect}.
    \item We evaluate these adaptations on the GLUE benchmark \citep{wang-etal-2018-glue} (\Cref{sec:GLUE}). This gives us results on how well the adaptations work for tasks that aim to evaluate general natural language understanding.
    \item We also evaluate the adaptations on the Visual Property Norms (VPN) \citep{hagstrom-johansson-2022-models} (\Cref{sec:VPN}). This gives us results on how well the adaptations work for zero-shot tasks that aim to evaluate visual conceptual knowledge in the models. 
    \item We compare the adapted VL models to their unimodal BERT-base \citep{devlin-etal-2019-bert} counterparts on the aforementioned evaluation tasks. The ensuing results should provide additional clarity on the natural language understanding of VL models.
    \item We also compare the adapted VL models to the multimodal FLAVA model \citep{singh2022flava} that requires no adaptation to text-only tasks.
\end{itemize}

%Another method instead of adapting multimodal models to a unimodal input is to already at the training stage move between multimodal and unimodal queries, to adapt the model to both types. We will not investigate this approach and only focus on adapting models already trained in a multimodal fashion.

%\section{Method}

%We evaluate seven different ways of adapting VL models to text-only input, described in \Cref{sec:adaptations}. The models we test these adaptations on are described in \Cref{sec:models} and our methods for evaluating the performance of the adaptations are described in \Cref{sec:evaluation}.

\begin{table*}[t]
    \centering
    \begin{tabular}{l c l l}
    \toprule
        Model & Size & Pre-train data & Backbone \\
        \midrule
        BERT-base & 110M & English Wiki, BookCorpus & - \\
        %BERT-large & 340M & English Wiki, BookCorpus \\
        FLAVA &  86M & CCNews, BookCorpus, PMD & - \\
        \hline
        CLIP-BERT & 110M & English Wiki, BookCorpus, CLIP-BERT V+L & CLIP \\
        LXMERT & 230M & LXMERT V+L & Faster R-CNN \\
        VisualBERT & 110M & English Wiki, BookCorpus, VisualBERT V+L & Faster R-CNN \\
    \bottomrule
    \end{tabular}
    \caption{The models evaluated with details on their pre-training data. The V+L datasets refer to model-specific VL datasets. With `FLAVA` we refer to the text encoder.}
    \label{tab:models}
\end{table*}

\section{Models}\label{sec:models}
We investigate adaptations to text-only input for the three multimodal models CLIP-BERT, LXMERT and VisualBERT. We also compare their results with those of a baseline BERT-base model and  FLAVA. The models are further described below and an overview of them can be found in \Cref{tab:models}.

For each of the multimodal models, we also describe how to make the model function without visual input. This is later used in some of the adaptations we evaluate, described in \Cref{sec:adaptations}.

All models evaluated in this work except for CLIP-BERT are provided by the Huggingface library \citep{wolf-etal-2020-transformers}. The pre-trained model weights for all models except for CLIP-BERT are also provided by this library. The CLIP-BERT weights are found in our public repository.

\subsection{VisualBERT}
VisualBERT is a single-stream model that has been initialized from pre-trained BERT-base weights and then further trained on MS COCO as well as VQA \citep{lin2014microsoft,balanced_vqa_v2}. As a result, it has been trained on 1.27M more texts and 0.12M more images than BERT-base. It utilizes a Faster R-CNN detector \citep{faster-r-cnn} as backbone, for which it has been trained on the features of the 36 first detections, meaning that it expects visual input features with shape $(36, 2048)$. 

\paragraph{Usage without visual input}
The single-stream architecture of this model implies that it simply concatenates the embeddings from the visual features with the word embeddings from the text input and then forwards this to the BERT encoder. 
Therefore, this model can be queried with text only without changing anything in the model architecture, since it simply means that only the word embeddings are fed to the BERT encoder. 

\subsection{LXMERT}
LXMERT is a dual-stream model trained on MS COCO, VQA, VG, GQA and VG-QA \citep{hudson2018gqa,zhu2016visual7w}. It has not been initialized from BERT-base weights. In total, it has been trained on 9.18M visual texts and 0.18M images. Similarly to VisualBERT, this model expects visual features of the shape $(36, 2048)$ from 36 Faster R-CNN detections.

\paragraph{Usage without visual input}
The dual-stream architecture of this model implies that it processes the visual embeddings and word embeddings in separate encoders before it fuses the information from them in a so called Cross-Modality Encoder. For this model it does not suffice to simply omit the visual input since it is expected by a separate visual encoder. However, the language output of the model is only affected by the visual input at a set of cross-attention sub-layers with residual connections in the Cross-Modality Encoder. Consequently, we can set the added residual from the cross-attention layer to zero and remove the visual encoder of the model.

\subsection{CLIP-BERT}
CLIP-BERT is a single-stream VL model that is architecturally very similar to VisualBERT. The main differences this model introduces are two, 1) it has a CLIP \citep{radford2021learning} backbone that generates visual features of dimension $(512,)$ for each image, and 2) it has been trained on 4.72M visual texts and 2.91M images, a vision-language dataset approximately four times larger than that of VisualBERT, in addition to having been initialized from BERT-base weights. 

\paragraph{Usage without visual input}
Similarly to VisualBERT, the single-stream architecture of this model implies that it can be queried with text only without changing anything in the model architecture, since it simply means that only the word embeddings are fed to the BERT encoder. 

\subsection{BERT-base}
Since all VL models we evaluate to some extent are based on BERT-base, we use this unimodal model as a baseline in our evaluations seen in \Cref{sec:evaluation}. 

We also create two additional baseline versions of BERT-base by further training the pre-trained model on LXMERT text data\footnote{The data is described in \url{https://github.com/airsplay/lxmert}.} and a subset of the English Wikipedia corpus from the Huggingface Datasets library \citep{lhoest-etal-2021-datasets} sampled to match the LXMERT text data in size, respectively. We do this to enable more fair comparisons to the evaluated VL models, since they have received additional training on text and images. These model versions are denoted by \texttt{trained-LXMERT} and \texttt{trained-Wikipedia}. The unchanged BERT-base model is denoted by \texttt{default}.

Since the original LXMERT model developed by \citet{tan-bansal-2019-lxmert} was not initialized from BERT weights, we also develop a third baseline version of BERT that has been trained from scratch on LXMERT text data for comparison. This model version is denoted by \texttt{trained-LXMERT-scratch}.

More information about the datasets used to train the BERT-base baselines and training procedures can be found in \Cref{app:datasets,app:training-procedure} respectively. 

\subsection{FLAVA}

FLAVA is a multimodal model that works for all combinations of VL modalities without any need for adaptation \citep{singh2022flava}. It sidesteps all issues related to the aforementioned VL models and can directly be evaluated for its linguistic capabilities. It consists of three separate parts: an image encoder, a text encoder, and a multimodal encoder that combines the input from the unimodal encoders. The unimodal encoders are pretrained on unimodal datasets and the full model is then trained end-to-end on the Public Multimodal Datasets (PMD) corpus \citep{singh2022flava}. We use the text encoder of this model as a baseline in our evaluations.

%Standard and further trained on LXMERT data as well as Wikipedia.

\section{Adaptations to text-only input}\label{sec:adaptations}
There are several ways to adapt a VL model to a text-only input. In this work we investigate and compare seven possible adaptations, as described below. Two of the adaptations described here have already been used for investigating the linguistic capabilities of VL models \citep{frank-etal-2021-vision,iki-aizawa-2021-effect}. Common for all adaptations is that their intended use is for evaluation of a pre-trained VL model (encoder) on text-only input. When we refer to the word \emph{adaptation} we refer to the adaptation of a VL model to text-only input.

The adaptions can be grouped into three different categories based on how they are implemented. For the first category, we simply remove the visual input to the VL model (\Cref{sec:adapt-text-only,sec:adapt-text-only-finetune}). For the second category, we provide the model with visual features that are constant and can be viewed as fillers, (\Cref{sec:adapt-avg,sec:adapt-black,sec:adapt-zeros,sec:adapt-tune}). For the third category, we provide the model with visual features predicted from text (\Cref{sec:adapt-predict}).

\subsection{Using model as-is without visual input}\label{sec:adapt-text-only}

All VL models considered in this work can be queried with text only, or after performing a small set of alterations to the model architecture without changing any pre-trained model weights, as described in \Cref{sec:models}. Thus, we can directly evaluate the pre-trained models on the text-only task of interest. This adaptation is denoted by \texttt{default}.

This adaptation is very simple to apply and does not require any additional computations, while it assumes that the VL model can be queried without visual input. It is also not certain that the models will function as intended due to the imposed train/test shift of this adaptation. To our knowledge, this approach has not been tested before.

\subsection{Fine-tuning model on text-only input}\label{sec:adapt-text-only-finetune}

 Before evaluating the pre-trained VL model we fine-tune it on a small text-only fine-tuning task, similarly to how several natural language understanding tasks are performed \citep{wang-etal-2018-glue,NEURIPS2019_4496bf24}. The idea is that this will acclimatize the model to the aforementioned domain shift. Similarly to the \texttt{default} adaptation, this also relies on being able to use the model  without visual input. 

We create two separate fine-tuning sets %in English 
for this adaptation. The sets have been extracted from the text part of the LXMERT training data and from English Wikipedia. Their sizes have been adapted to match those of typical fine-tuning sets for e.g. SuperGLUE \citep{NEURIPS2019_4496bf24} and we have ensured that the number of tokens in each fine-tuning set is roughly equal. More information about the datasets can be found in \Cref{app:datasets}.

Finetuning the VL models on each of these sets should give us results on both the performance of the method, and on how dependent it is on the chosen fine-tuning set. %We expect the Wikipedia fine-tuning to be more beneficial for the GLUE evaluation and vice versa for the LXMERT set for the Visual Property Norms evaluation. 
These adaptations are denoted by \texttt{no-visual-features-finetuned\-LXMERT} and \texttt{no-visual-features-fine\-tuned-Wikipedia} respectively.

This method avoids having to work with image feature extractors and image data. However, it requires setting up a training algorithm and additional computations. Moreover, since the full model needs to be trained for this adaptation, it is more sensitive to hyperparameter choices. It is also not certain whether it is sufficient to perform fine-tuning on text to acclimatize VL models to a text-only input. More information on tuning procedures can be found in \Cref{app:training-procedure}.

\subsection{Using averaged visual features from the training dataset}\label{sec:adapt-avg}

In this method we give the VL model a constant visual feature input together with the text of interest at evaluation, where the visual features are the average of all the visual features in the train or evaluation data of the model. The provided visual-features should then be kept in-distribution, while they also are uninformative. 
This adaptation has already been used by \citet{frank-etal-2021-vision} for ablating visual input to VL models. We denote it by \texttt{avg-visual-features}.

No assumptions or changes to the model architecture are necessary for this method. However, it requires access to the datasets used to train the model of interest and the computation of the averaged visual feature vector.

We calculate the averaged visual feature vector for CLIP-BERT based on the CLIP features of its training data. We also calculate the averaged visual features and position vectors for LXMERT from its corresponding training data. We take the average across training samples per detection for the LXMERT visual features such that we get one average feature vector for the first detection, another for the second detection and so forth up to the 36th.%, since this slightly outperforms taking the average across training samples and detections.

The original released VisualBERT visual features are not compatible with the Huggingface implementation of the model used in this work. %Since the Huggingface implementation of VisualBERT is seemingly based on the same image detection model as LXMERT with the same shape of the features, 
We instead provide VisualBERT with the LXMERT averaged visual features, since they are compatible.% This is not optimal, since VisualBERT should receive averaged VisualBERT features. It was however a necessary step to take since the original released VisualBERT visual features are not compatible with the Huggingface implementation of the model. 

\subsection{Using visual features from a black image}\label{sec:adapt-black}

The idea is yet again to give the VL model a constant visual feature input together with the text of interest. In this case, the visual features are extracted from a black image using the model backbone. The model then receives a visual input similar to what it has been trained on, while it does not contain any information. This adaptation has already been used by \citet{iki-aizawa-2021-effect} for evaluating e.g. VisualBERT and LXMERT on GLUE. We denote it by \texttt{zero-image-visual-features}.

Similarly to the averaged features adaptation, this adaptation makes no assumptions about the model of interest. However, it requires access to the backbone of the model and the computation of the visual features from a black image.

We use the LXMERT feature extractor to extract 36 detections with their visual features and bounding boxes from a black image. The extractor is a Faster R-CNN model developed by \citet{faster-r-cnn}. These features are then given to LXMERT and VisualBERT during evaluation. For CLIP-BERT we use CLIP to extract visual features from the same black image.

\subsection{Using constant zero vector visual features}\label{sec:adapt-zeros}

We give the model of interest constant visual features, and the positions of bounding boxes in the case of LXMERT, that are zeros. There are no guarantees that this method will work well for adapting VL models to a text-only input. It is however easy to implement and can be seen as a baseline to be compared with the other adaptations. To the knowledge of the authors, this method has not been used previously. We denote it by \texttt{zeroed-visual-features}.

\subsection{Using tuned visual features}\label{sec:adapt-tune}

We tune the visual features to a frozen version of the model of interest, and then use these constant features at evaluation together with the text of interest. To the knowledge of the authors, this method has not been used previously to adapt VL models to a text-only input. However, the key idea of tuning the input to the model has been used in previous works \citep{qin-eisner-2021-learning,tsimpoukelli2021multimodal}.

%on e.g. prompt engineering \citep{qin-eisner-2021-learning} or tuning backbones to VL models \citep{tsimpoukelli2021multimodal}.

We tune visual features to frozen and pre-trained versions of CLIP-BERT, VisualBERT and LXMERT respectively. We tune on the same LXMERT and Wikipedia sets used for the adaptation described in \Cref{sec:adapt-text-only-finetune}. More information on tuning procedures can be found in \Cref{app:training-procedure}.

This method offers more flexibility for finding the most suitable constant visual features for a VL model evaluated on text-only tasks. However, it also requires setting up the training, more computations and is more sensitive to hyperparameter tuning. We denote these adaptations on the different fine-tuning sets by \texttt{finetuned-LXMERT-visual-features} and \texttt{finetuned-Wikipedia-visual\-features} respectively.

\subsection{Predicting visual features from text}\label{sec:adapt-predict}

Some feature extractors map text representations and visual representations to the same parametric space. Consequently, they can be used to ``imagine'' visual features from text. The CLIP model serving as a backbone for the CLIP-BERT model has this capability and can be used to generate visual features from text during evaluation on text-only tasks. We implement it for the CLIP-BERT model and denote it by \texttt{imagined-visual-features}.

This method is quite simple to implement, while it requires access to CLIP and computing the visual features from the evaluation corpus. It is also not clear how well CLIP representations work for text that is not specifically related to visual concepts.

%\subsection{Train model from scratch on both unimodal and multimodal input}

%This method is not one of the seven adaptations investigated by us in this work since we only focus on already pre-trained VL models. Nonetheless it still is worth mentioning this method since it potentially is one of the most promising ways to adapt a model to both unimodal and multimodal input. 

\section{Evaluation methods}\label{sec:evaluation}
%To assess the performance of our text-only adaptations, we perform task-based-evaluations on the two evaluation tasks GLUE and Visual Property Norms, further described in \Cref{sec:GLUE,sec:VPN}. 

To assess the performance of our text-only adaptations, we firstly evaluate them on the GLUE benchmark, described in \Cref{sec:GLUE}. These evaluations will give us results on how well the models and their adaptations work for general natural language understanding tasks. This benchmark has been used by both \citet{devlin-etal-2019-bert} and \citet{iki-aizawa-2021-effect} to evaluate natural language understanding capabilities of BERT and VL models. 

Furthermore, to assess the performance of the adaptations on text domains that are more focused on visual concepts, we perform evaluations on VPN, further described in \Cref{sec:VPN}. This will provide us with results on tasks the VL models potentially are more attuned to, complementing the general GLUE results. 

\begin{table}[h]
    \centering
    \begin{tabular}{l l}
    \toprule
        MLM query & Gold labels \\
        \midrule
        a cow usually is [...]  & black, white \\ 
        a mug has a [...] & handle \\ 
        q: a greeting card has? a: [...] & pictures \\
    \bottomrule
    \end{tabular}
    \caption{Query samples from the VPN dataset for the concepts \emph{cow}, \emph{mug} and \emph{greeting card} using three out of nine possible query templates. The [...] is typically replaced with a \texttt{[MASK]} token.}
    \label{tab:VPN-examples}
\end{table}

\subsection{GLUE}\label{sec:GLUE}
The General Language Understanding Evaluation (GLUE) benchmark has the aim to evaluate model performance across several NLU tasks. It was developed by \citet{wang-etal-2018-glue} and has since then been used to evaluate the natural language understanding of several LMs, including BERT.

GLUE contains nine different tasks testing for grammatical correctness understanding (CoLA), sentiment classification (SST-2), semantic equivalence detection on different text domains (MRPC, QQP, STS-B), textual entailment (MNLI, RTE), answer extraction from text (QNLI) and reading comprehension (WNLI). All tasks are sentence classification tasks and have corresponding train and validation sets for fine-tuning. %Models fine-tuned on these tasks can then be evaluated on held-out test sets.

VL models have already been evaluated on GLUE by \citet{iki-aizawa-2021-effect} using the black image adaptation method listed in \Cref{sec:adapt-black}. We extend the GLUE evaluation to include all alternative adaptation approaches listed in \Cref{sec:adaptations}.

To evaluate the performance of our adaptations on GLUE, we first fine-tune our selected multimodal models with each adaptation on the training sets of the GLUE tasks. We then report the validation scores of the models and their adaptations. More information about the fine-tuning procedures can be found in \Cref{app:training-procedure}.

\begin{figure*}[t]
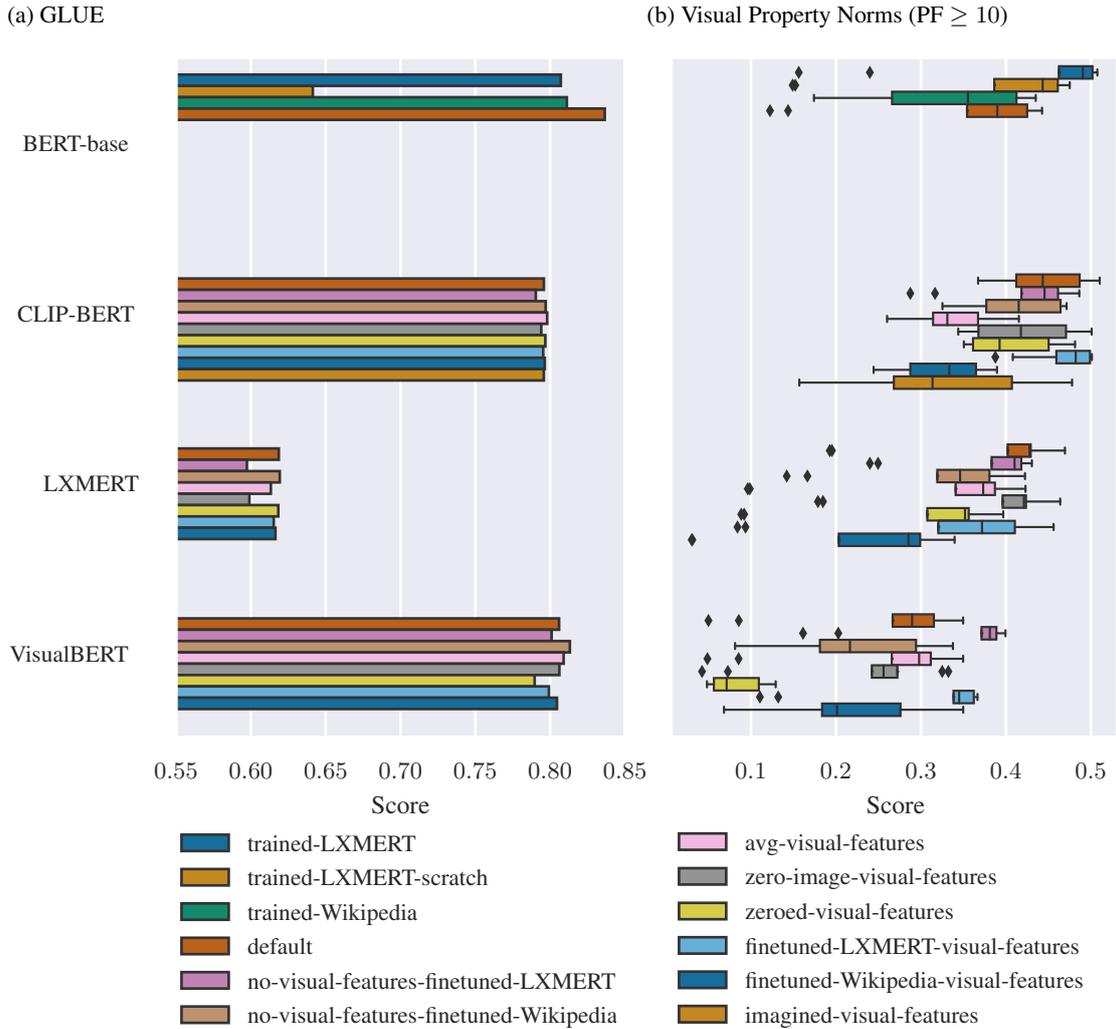

    \begin{center}
    \begin{subfigure}[b]{0.52\textwidth}
    \caption{GLUE}
    \label{fig:glue-results}
    \begin{adjustbox}{clip,trim=0cm 0cm 7.3cm 0.1cm}
    \hspace{-0.7cm}\input{figures/GLUE_results.pgf}
    \end{adjustbox}
    \end{subfigure}
    \begin{subfigure}[b]{0.38\textwidth}
    \caption{Visual Property Norms (PF $\geq10$)}
    \label{fig:vpn-results}
    % left, bottom, right, and top
    \begin{adjustbox}{clip,trim=2.6cm 0cm 7.8cm 0.1cm}
    \input{figures/visual_property_norms_results.pgf}
    \end{adjustbox}
    \end{subfigure}
    
    % legend
    \vspace{-0.2cm}
    \hspace{2.2cm}
    \begin{adjustbox}{clip,trim=8.5cm 7.65cm 1.6cm 0.3cm}
    \hspace{-0.7cm}\input{figures/GLUE_results.pgf}
    \end{adjustbox}
    \hspace{0.4cm}
    \begin{adjustbox}{clip,trim=8.5cm 4.9cm 1.6cm 3.1cm}
    \hspace{-0.5cm}\input{figures/GLUE_results.pgf}
    \end{adjustbox}
    \end{center}
    \caption{Results on GLUE and VPN from evaluating different adaptations to text-only input. The GLUE results are given by the mean of the scores for the development sets of all tasks, excluding WNLI. The metric used for each task is F1 score for QQP and MRPC, Matthews correlation for CoLA, Spearman correlation for STS-B, and accuracy for the remaining tasks. For VPN the box length indicates prompt sensitivity over nine different query templates.}
    \label{fig:results}
\end{figure*}

\begin{table*}[h]
    \centering
    \begin{tabular}{lrrrrrrrrr}
    \toprule
     &  CoLA & MNLI & MRPC & QNLI &  QQP &  RTE & SST-2 & STS-B \\
    \midrule
    BERT-base  &                 61.1 &     84.6 &     87.3/91.2 &     91.9 &     91.1/88.0 &     70.4 &     93.7 &      88.2 \\
    FLAVA      &                 50.1 &     81.6 &     83.6/88.3 &     87.8 &     90.4/87.2 &     55.6 &     92.4 &      87.1 \\
    \hline
    CLIP-BERT  &                 55.4 &     83.2 &     75.5/84.1 &     89.8 &     91.1/88.0 &     58.1 &     92.0 &      87.8 \\
    LXMERT     &                 15.9 &     68.1 &     69.9/81.6 &     68.0 &     84.1/76.8 &     58.5 &     86.6 &      40.1 \\
    VisualBERT &                 53.3 &     83.7 &     80.4/86.4 &     90.7 &     90.9/87.6 &     67.5 &     91.7 &      89.6 \\
    \bottomrule
    \end{tabular}
    \caption{GLUE development set results per task for the best performing adaptation on average. The best performing adaptation for each model is `default` for BERT-base, `avg-visual-features` for CLIP-BERT, `no-visual-features-finetuned-Wikipedia` for LXMERT and `no-visual-features-finetuned-Wikipedia` for VisualBERT . We report Matthew's correlation for CoLA, average accuracy for MNLI, accuracy/F1 score for MRPC, accuracy for QNLI, accuracy/F1 for QQP and accuracy for RTE and SST-2 and Spearman correlation for STS-B.}
    \label{tab:GLUE_task_results}
\end{table*}

\begin{table}[h]
    \centering
    \begin{tabular}{ll}
    \toprule
     &  VPN Score \\
    \midrule
    BERT-base  &              49.1 $\pm$ 13.2 \\
    FLAVA &            30.7 $\pm$ 6.9 \\
    \hline
    CLIP-BERT  &              48.2 $\pm$  4.2 \\
    LXMERT &              42.8 $\pm$ 10.8 \\
    VisualBERT &              38.1 $\pm$ 9.1 \\
    \bottomrule
    \end{tabular}
    \caption{VPN results for the best \mbox{performing adap}\-tations. The best performing \mbox{adaptation for each} \mbox{model is `trained-LXMERT' for BERT-base, `finetuned-} \mbox{LXMERT-visual-features' for CLIP-BERT, `default' for} \mbox{LXMERT and `no-visual-features-finetuned-LXMERT'} \mbox{for VisualBERT. We report the results as median $\pm$ stan}\-\mbox{dard deviation over the nine query templates.}}
    \label{tab:VPN_task_results}
\end{table}

\subsection{Visual Property Norms}\label{sec:VPN}
Our current VL models are not necessarily the best fit for general NLU tasks such as GLUE \citep{iki-aizawa-2021-effect,yun-etal-2021-vision-language}. Therefore, we also evaluate them on a task we assume they are more suitable to. Visual Property Norms (VPN) essentially queries a model for the basic visual properties of a set of concepts \citep{hagstrom-johansson-2022-models}. It is a text-only task, while it explicitly focuses on visual properties and concepts. Thus, if the VL models should perform particularly well on any text-only task, this would be the one. \Cref{tab:VPN-examples} displays examples of queries from the VPN dataset.   

The VPN dataset is a zero-shot evaluation task that evaluates a model using masked language modelling (MLM), an objective our models already have been trained on. To mitigate issues with prompt-sensitivity of LMs, nine different query templates are applied during evaluation.  

VPN is a version of the CSLB concept property norms dataset \citep{devereux2014centre} filtered to only contain visual conceptual features. The original property norms dataset was created with the help of 123 human participants asked to list the features of a set of concepts. Each concept has in total been exposed to 30 humans and the maximum frequency of a feature reported for a concept is then 30 and the minimum 2. This frequency is referred to as Production Frequency (PF). 

VPN has been segmented into five partitions based on thresholding of PFs. We evaluate our adaptations on the segment for which PF $\geq 10$, such that ten or more annotators jointly have produced the visual features in this set. It consists of 2,001 feature entries for 621 different concepts.% such as \emph{apple}, \emph{boots}, \emph{tiger} and \emph{wine}.

\section{Results}
We report the evaluation results for CLIP-BERT, LXMERT and VisualBERT with the seven potential adaptions to text-only input in \Cref{fig:results}. We also report the results for our four BERT-base baselines. For GLUE we report the macro-averaged score over the GLUE tasks. The score for each task is measured using its corresponding predefined metric described by \citet{wang-etal-2018-glue}. For VPN evaluation we report the mean average precision (mAP) score averaged over each concept and relation per query template. 

We also report the evaluation results on GLUE for each task and model for the best performing adaptation measured by average GLUE score in \Cref{tab:GLUE_task_results}. \Cref{tab:VPN_task_results} similarly reports the model scores on VPN for the best adaptation on average. We compare these results to those of the FLAVA text encoder that requires no adaptation. Complete numerical results can be found in \Cref{app:results}.

We format our discussion around a set of statements that can be made with respect to the results of this work, as follows.

\paragraph{Model performance on GLUE is more sensitive to pre-training than to adaptation}
Model performance on GLUE varies insignificantly between different adaptations for each model in \Cref{fig:glue-results}. The CLIP-BERT performance varies with less than 0.01 score points between adaptations and the VisualBERT performance with at most 0.02 score points between 'no-visual-features-finetuned-Wikipedia' and 'zeroed-visual-features'. The LXMERT performance also has a performance difference of at most 0.02 score points between 'default' and 'zero-image-visual-features'. 

The largest performance difference on the GLUE benchmark can be observed between models, where LXMERT and BERT-base trained from scratch on LXMERT data perform significantly worse in comparison to the other models. Most likely, this is due to that the models were not initialized from BERT weights and consequently were not tuned to more general language usage.

A possible explanation for why the adaptation methods seem to matter so little for the GLUE results is that the benchmark is not zero-shot. The fine-tuning performed on each task might provide a sufficient of signal to the model for it to adapt to the unimodal domain.

Lastly, our results on GLUE for LXMERT and VisualBERT differ from those obtained by \citet{iki-aizawa-2021-effect}. Especially with respect to LXMERT for which we observe a significant performance difference compared to the other VL models, while the same cannot be observed in the results by \citet{iki-aizawa-2021-effect}. However, this should not raise any concerns about the robustness of the results, since we evaluated the original released models, while \citet{iki-aizawa-2021-effect} evaluated models that had been unified and trained on the same data by \citet{bugliarello-etal-2021-multimodal}. And as we observed previously, GLUE results are sensitive to pre-training process.

\paragraph{Performance on Visual Property Norms is sensitive to adaptation}

In contrast to the observations made for GLUE, performance on VPN differs significantly between different adaptions in \Cref{fig:vpn-results}. In contrast to GLUE, this task is zero-shot and may provide a greater challenge for models that are not sufficiently tuned to the unimodal text domain.

Additionally, it is worth noting that the model performance is quite sensitive to the choice of query template. However, this is not entirely unexpected since it has been shown that LMs are prompt-sensitive in prompt-based retrieval evaluations \citep{cao-etal-2021-knowledgeable,jiang2020know}.

\paragraph{Different adaptations perform differently for different models on Visual Property Norms}

For CLIP-BERT, the most suitable adaptation for evaluation on VPN is to provide the model with visual features that have been tuned on LXMERT text data. For LXMERT, the best approach is to use the model as-is without visual input, and for VisualBERT the best adaptation is to fine-tune the model on LXMERT data without visual features. Common for all of these adaptations is that they involve some kind of prior tuning on LXMERT text data. A potential explanation for this is that the LXMERT data is more similar to VPN and results in a smaller domain shift.

%The models also differ in what adaptations work worst. For CLIP-BERT this is imagined visual features, for LXMERT it is visual features fine-tuned on Wikipedia and for VisualBERT it is zeroed visual features.

An explanation for the varying adaptation fits between VL models is potentially found by looking at the different pre-training datasets and architectures of the models. VisualBERT has been tuned on much less data compared to the other models, and may therefore benefit from more training in general. Additionally, the single-stream CLIP-BERT and VisualBERT models process all linguistic and visual information in a joint manner, without the same ability to disentangle signals as the dual-stream LXMERT model.

%Additionally, LXMERT is a dual-stream architecture for which visual information impacts the linguistic representations in a more late-fusion fashion, through a limited set of residual connections in the Cross-Modality Encoder. In contrast to this, the single-stream CLIP-BERT and VisualBERT models process all linguistic and visual information in a joint manner, without the same ability to disjoin signals. %These models may therefore respond differently to the adaptations.

%\vspace{-4mm}
\paragraph{FLAVA does not outperform adapted VL models}

On both GLUE and especially VPN, FLAVA is not better than the adapted VL models. This contrasts the GLUE results reported for FLAVA and other VL models by \citet{singh2022flava}. The difference in results may arise from differences in fine-tuning methods for GLUE and that we do not evaluate unified VL models.

%that FLAVA previously has been compared to unified VL models fine-tuned differently on GLUE.  

Based on our results, adapting VL models to text-only input works better or equal to developing a model to work for all modalities from the start, as was done for FLAVA. However, since all models evaluated have been trained on different datasets with different objectives we cannot draw certain conclusions related to model design.

\paragraph{BERT-base baselines outperform vision-and-language models regardless of adaptation}

Lastly, we can observe in \Cref{fig:results} how default BERT-base and BERT-base trained on LXMERT text data each have among the best performances on GLUE and VPN respectively. For GLUE it is expected: \citet{iki-aizawa-2021-effect} have already observed that VL models on average perform worse on GLUE. \citet{yun-etal-2021-vision-language} also found similar results when they compared the quality of the linguistic representations of VisualBERT to those of BERT-base. The general natural language understanding capabilities evaluated in GLUE are potentially not easy to learn from a visual modality, explaining why the VL models did not perform better on this task. Our results on VPN are perhaps more surprising. 

From their visual training, the VL models should more easily have gained natural language understanding capabilities necessary for better performance on VPN. Three potential explanations for why a LM still outperforms a VL model on VPN are 1) BERT-base is better tuned and therefore has a better overall performance, 2) the VL models evaluated in this work do not learn more about visual concepts from images compared to text that has been curated to contain visual information, or 3) the VPN task does not accurately measure the visual conceptual information we have in mind. More investigations are necessary to accurately determine the reason. In support of explanation (2), \citet{abdou-etal-2021-language} found that there are similarities between color representations in LMs and actual perceptual color spaces, indicating that visual perceptual information may be found in text. 

%Our results indicate that better natural language understanding is not gained by default from multimodal training. However, none of the models evaluated in this work were developed with the goal of achieving better natural language understanding by multimodal training. This potentially explains some of our results, and provides an interesting avenue for future research in developing models that have competitive performance on both unimodal and multimodal tasks.

We should also note that none of the models evaluated in this work were developed with the goal of achieving better natural language understanding by multimodal training. This potentially explains some of our results, and provides an interesting avenue for future research in developing models that have a better performance on both unimodal and multimodal tasks.
 
 %it is perhaps not surprising that our current VL models This is perhaps not surprising since all models were developed with the aim of achieving good results on e.g. VQA and captioning tasks. That is, none of the models were developed with the goal of achieving better natural language understanding by multimodal training. Future work that investigates this avenue by e.g. training models to have competitive performance on both unimodal and multimodal input would be interesting.

%These results do not mean that the idea of having models learn language from more than text has failed. They do however indicate that there is more work to be done on developing models that use multimodal pre-training to improve on their natural language understanding.

\section{Related Work}
%What to put here? Extract references from other parts of article? Look at discussed previous works!

As previously mentioned, \citet{iki-aizawa-2021-effect} have already looked at the language understanding capabilities of VL models, while they only looked at one way of adapting these models to a text-only input and only evaluated on GLUE. \citet{yun-etal-2021-vision-language} also look at the language understanding capabilities of VL models by evaluating the linguistic representations of VisualBERT and compare them to a BERT-base model that has been trained on the same text data. In contrast to their work, we investigate several VL models and evaluate their performance on language generation tasks.

\citet{bugliarello-etal-2021-multimodal}, \citet{hessel-lee-2020-multimodal}, \citet{thrush2022winoground} and \citet{frank-etal-2021-vision} also perform extensive evaluations of several VL models such as LXMERT and VisualBERT. In contrast to our work, they primarily focus on the VL performance of the models, and do not consider the model performance on text-only input. 

\citet{tan-bansal-2020-vokenization} introduce a new method for enriching the textual representations of a model by training on visual information. Their method results in a model that can be directly applied to text-only tasks and outperforms its standard BERT model counterpart on GLUE. %They showed that improved model performance could be reached on GLUE tasks by training a BERT model on an additional voken classification task. 
This method provides a parallel research avenue compared to adapting VL models to text-only input.

%A VL model that has showed promise on text-only tasks is the FLAVA model. It was developed by \citet{Singh_2022_CVPR} to perform well on vision tasks, language tasks and vision-and-language tasks simultaneously. Compared to the models investigated in our work, the FLAVA model shows comprehensive performance improvements from multimodal training, also compared to unimodal model counterparts.  

%Work by Iki et al.

%Work by VL or vision-for-language

%FrozenNN

% Limitations
% * Extent to which GLUE evaluates general natural language understanding?
% * Models not unified

% FLAVA bad results explanation
% * Text model separate and thus not influenced as much by images?
% * Greater domain shift between training and GLUE+VPN?

\section{Conclusions}

We have investigated and compared seven possible adaptations of CLIP-BERT, LXMERT and VisualBERT to text-only input by evaluation on GLUE and Visual Property Norms. We can conclude that care should be put into adapting these pre-trained VL models to text-only input for better performance on zero-shot tasks, while the choice of adaptation method seems to be less impactful on tasks coupled with fine-tuning sets.

%Finally, we have observed that a unimodal LM has a performance that is better or comparative to that of VL models on text-only tasks, regardless of how these models were adapted to text-only input. At the same time, other works have introduced the voken classification training method and the FLAVA model, approaches that inherently can handle purely linguistic signals and display a performance that is generally better compared to that of their unimodal counterparts. This indicates that better natural language understanding is not gained by default from multimodal training and that .   offering promising alternatives to the VL models and text-only adaptations investigated in this work. 

Finally, we have observed that a unimodal LM has a performance on text-only tasks that is better or comparative to that of its VL model counterparts, regardless of how these counterparts were adapted to text-only input. Seemingly, improved pure text capabilities are not guaranteed from simply training a model on arbitrary multimodal tasks. This agrees with and solidifies previous research results on VL models.

\section*{Acknowledgements}

We would like to thank the reviewers for their valuable feedback and advice on related work.

This work was partially supported by the Wallenberg AI, Autonomous Systems and Software Program (WASP) funded by the Knut and Alice Wallenberg Foundation.
%This work was supported by the projects \emph{Interpreting and Grounding Pre-trained Representations for NLP} and \emph{Representation Learning for Conversational AI}, both funded by Wallenberg  AI,  Autonomous  Systems  and  Software Program (WASP) funded  by  the  Knut  and  Alice Wallenberg Foundation. 
The computations were enabled by resources provided by the Swedish National Infrastructure for Computing (SNIC), partially funded by the Swedish Research Council through grant agreement no. 2018-05973.

\clearpage
%\section*{Acknowledgements}

\bibliography{anthology,custom}
\bibliographystyle{acl_natbib}

\appendix

\section{Datasets for training and tuning}\label{app:datasets}

More detailed information on the datasets used for training and fine-tuning the models investigated in this work can be found here.

\subsection{Training of BERT baselines}

More information about the LXMERT and Wikipedia training datasets used for the BERT-base baselines can be found in \Cref{tab:bert-sets}. By training the BERT model on LXMERT text data, it will have seen the same textual information as LXMERT. And by training it on the Wikipedia data, it will have seen the same amount of text as LXMERT.

\begin{table*}[h]
    \centering
    \begin{tabular}{llccc}
    \toprule
       Corpus & Partition & \# of samples & \# of tokens & \# of tokens/sample \\
       \midrule
        LXMERT & train & 9.0M & 59.0M & 6.6 \\
        & dev & 0.2M & 1.4M & 6.8 \\
        Wikipedia & train & 4.4M & 59.0M & 13.4 \\
        & dev & 0.1M & 1.3M & 13.4 \\
    \bottomrule
    \end{tabular}
    \caption{The two text datasets used for developing two additional BERT-base baselines. The number of samples are roughly equal to the number of sentences for these datasets. The LXMERT data is the text part of the LXMERT training data. Wikipedia is a subset of general English Wikipedia texts that has been adapted to match the LXMERT data in total number of tokens.}
    \label{tab:bert-sets}
\end{table*}

\subsection{Fine-tuning model on text-only input}

The LXMERT and Wikipedia datasets used for fine-tuning text-only versions of VL models are further described in \Cref{tab:finetune-sets}. The two fine-tuning sets cover quite different domains. This is already visible from the tokens/sample count in the table, in which the Wikipedia corpus generally contains long sentences and the LXMERT corpus generally contains shorter sentences more suitable for image captions.

\begin{table*}[h]
    \centering
    \begin{tabular}{llccc}
    \toprule
       Corpus & Partition & \# of samples & \# of tokens & \# of tokens/sample \\
       \midrule
        LXMERT-f & train & 9,500 & 63,000 & 6.6 \\
        & dev & 3,300 & 22,000 & 6.6 \\
        Wikipedia-f & train & 4,600 & 63,000 & 13.7 \\
        & dev & 1,600 & 22,000 & 13.5 \\
    \bottomrule
    \end{tabular}
    \caption{The two text datasets used for fine-tuning, denoted by the ``-f'' ending. The number of samples are roughly equal to the number of sentences for these datasets.}
    \label{tab:finetune-sets}
\end{table*}

\section{Training procedures}\label{app:training-procedure}

More detailed information on our training procedures can be found here.

\subsection{Training BERT-base baselines}
For the training of BERT-base on both the LXMERT and Wikipedia datasets we use an MLM objective, a batch size of 16,384 and learning rate \SI{5e-5} until the model performance on the dev set had converged. The maximum training time was at most 23 hours on 32 Tesla T4 GPUs.

\subsection{Fine-tuning model on text-only input}

We fine-tune the models using an MLM objective, batch size of 256 and learning rate \SI{5e-5} until the model performance on the dev set had converged. The maximum training time was two hours on eight Tesla T4 GPUs.

\subsection{Using tuned visual features}

We tune the visual features using an MLM objective, batch size 64 and a learning rate of 0.05 until the model performance had converged on the dev set. The maximum training time was 18 hours on one Tesla T4 GPU.

\subsection{Fine-tuning on GLUE}
For the GLUE fine-tuning, we tune our models for four epochs with a learning rate of \SI{3e-5}, weight decay of 0.01 and batch size of 32. The longest tuning time was four hours on two A100 GPUs. We then pick the model checkpoint with the best validation score during training for evaluation.

\section{Complete numerical results}
\label{app:results}

The complete numerical results on GLUE and VPN can be viewed in \Cref{tab:full-glue-results,tab:full-vpn-results} respectively.

\begin{table*}[]
    \centering
    \begin{tabular}{lll}
\toprule
Model & Adaptation & Score          \\
\midrule
BERT-base & trained-LXMERT &   80.7 \\
           & trained-LXMERT-scratch &   64.1 \\
           & trained-Wikipedia &   81.1 \\
           & default &   83.7 \\
FLAVA & default &   78.8 \\
\midrule
CLIP-BERT & default &   79.6 \\
           & no-visual-features-finetuned-LXMERT &   79.0 \\
           & no-visual-features-finetuned-Wikipedia &   79.7 \\
           & avg-visual-features &   79.8 \\
           & zero-image-visual-features &   79.4 \\
           & zeroed-visual-features &   79.7 \\
           & finetuned-LXMERT-visual-features &   79.5 \\
           & finetuned-Wikipedia-visual-features &   79.6 \\
           & imagined-visual-features &   79.6 \\
LXMERT & default &   61.9 \\
           & no-visual-features-finetuned-LXMERT &   59.7 \\
           & no-visual-features-finetuned-Wikipedia &   61.9 \\
           & avg-visual-features &   61.3 \\
           & zero-image-visual-features &   59.9 \\
           & zeroed-visual-features &   61.8 \\
           & finetuned-LXMERT-visual-features &   61.5 \\
           & finetuned-Wikipedia-visual-features &   61.6 \\
VisualBERT & default &   80.6 \\
           & no-visual-features-finetuned-LXMERT &   80.1 \\
           & no-visual-features-finetuned-Wikipedia &   81.3 \\
           & avg-visual-features &   80.9 \\
           & zero-image-visual-features &   80.6 \\
           & zeroed-visual-features &   79.0 \\
           & finetuned-LXMERT-visual-features &   79.9 \\
           & finetuned-Wikipedia-visual-features &   80.5 \\
\bottomrule
\end{tabular}
    \caption{The adaptation and baseline results for GLUE seen in \Cref{fig:glue-results}.}
    \label{tab:full-glue-results}
\end{table*}

\begin{table*}[]
    \centering
    \begin{tabular}{llrr}
\toprule
           &                                     & \multicolumn{2}{l}{Score} \\         
Model & Adaptation & Median &  Standard deviation \\
\midrule
BERT-base & trained-LXMERT &              49.1 & 13.2 \\
           & trained-LXMERT-scratch &              44.3 & 13.2 \\
           & trained-Wikipedia &              35.5 &  9.5 \\
           & default &              39.0 & 12.4 \\
FLAVA & default &              30.7 &  6.9 \\
\midrule
CLIP-BERT & default &              44.3 &  4.7 \\
           & no-visual-features-finetuned-LXMERT &              44.5 &  7.0 \\
           & no-visual-features-finetuned-Wikipedia &              41.5 &  5.4 \\
           & avg-visual-features &              33.1 &  5.0 \\
           & zero-image-visual-features &              41.8 &  5.8 \\
           & zeroed-visual-features &              39.3 &  4.9 \\
           & finetuned-LXMERT-visual-features &              48.2 &  4.2 \\
           & finetuned-Wikipedia-visual-features &              33.3 &  4.8 \\
           & imagined-visual-features &              31.4 & 10.2 \\
LXMERT & default &              42.8 & 10.8 \\
           & no-visual-features-finetuned-LXMERT &              41.0 &  7.5 \\
           & no-visual-features-finetuned-Wikipedia &              34.6 & 10.0 \\
           & avg-visual-features &              37.3 & 12.9 \\
           & zero-image-visual-features &              42.1 & 11.0 \\
           & zeroed-visual-features &              35.2 & 12.0 \\
           & finetuned-LXMERT-visual-features &              37.2 & 13.9 \\
           & finetuned-Wikipedia-visual-features &              28.5 & 11.7 \\
VisualBERT & default &              29.0 & 10.9 \\
           & no-visual-features-finetuned-LXMERT &              38.1 &  9.1 \\
           & no-visual-features-finetuned-Wikipedia &              21.6 &  9.0 \\
           & avg-visual-features &              29.8 & 11.0 \\
           & zero-image-visual-features &              25.6 & 10.2 \\
           & zeroed-visual-features &               7.1 &  3.1 \\
           & finetuned-LXMERT-visual-features &              34.5 & 10.3 \\
           & finetuned-Wikipedia-visual-features &              20.1 &  9.9 \\
\bottomrule
\end{tabular}
    \caption{The adaptation and baseline results for VPN.}
    \label{tab:full-vpn-results}
\end{table*}

\end{document}